\documentclass{nst}

\usepackage{subfigure,dcolumn}
\usepackage{epstopdf}
\usepackage[version=4]{mhchem}
\usepackage{upgreek}
\usepackage{hyperref}

\usepackage{algorithm}
\usepackage{algorithmic}

\begin{document}

\title{Correcting Sensor-Induced Distribution Drift with Wasserstein Adversarial Learning}

\author{Saraa Ali}
\email[Corresponding author]{thraaali@hse.ru}
\affiliation{Laboratory of Methods for Big Data Analysis, HSE University, Pokrovsky Boulevard 11, Moscow 109028, Russia}

\author{Vladimir Bocharnikov}
\affiliation{Laboratory of Methods for Big Data Analysis, HSE University, Pokrovsky Boulevard 11, Moscow 109028, Russia}

\author{Fedor Ratnikov}
\affiliation{Laboratory of Methods for Big Data Analysis, HSE University, Pokrovsky Boulevard 11, Moscow 109028, Russia}

\author{Mikhail Hushchyn}
\affiliation{Laboratory of Methods for Big Data Analysis, HSE University, Pokrovsky Boulevard 11, Moscow 109028, Russia}

\author{Artem Ryzhikov}
\affiliation{Laboratory of Methods for Big Data Analysis, HSE University, Pokrovsky Boulevard 11, Moscow 109028, Russia}

\author{Denis Derkach}
\affiliation{Laboratory of Methods for Big Data Analysis, HSE University, Pokrovsky Boulevard 11, Moscow 109028, Russia}

\begin{abstract}
The quality of recorded data depends on the stability of the sensor system that acquires it. Sensor motion and aging can degrade the performance and stability of downstream data-driven methods.
We present a Wasserstein-GAN-inspired approach for unsupervised inference of physically interpretable transformation parameters that map a changed detector response distribution back to a nominal reference distribution. In contrast to standard generative modeling, the generator is used as a learnable calibration transformation whose trainable weights represent the sought parameters, while the critic provides a distributional distance signal via the Wasserstein objective. We validate the approach on a tracking-detector toy model with controlled layer shifts and demonstrate its application on high-granularity Geant4-simulated calorimeter data with cell-wise aging effects. The method recovers aging coefficients for individual cells with correlation to ground truth and improves agreement between calibrated and reference energy-sum distributions, while exhibiting the expected degradation at increasing channel-to-channel noise levels. These results indicate that adversarial distribution matching can serve as a data-driven component of calibration strategies in settings where direct labels for degradation parameters are unavailable.
\end{abstract}


\maketitle

\section{Introduction}\label{sec1}

The integrity of any data-driven analysis pipeline depends on the quality and stability of the measurements provided by the underlying sensor system \cite{ABDULAI2025145101}. In experimental environments, sensor responses are rarely stationary: mechanical drifts, environmental conditions, and long term material aging can introduce systematic distortions that accumulate over time \cite{STAHMANN2025111236}. Even when these effects are small at the level of individual channels, they can propagate to downstream reconstruction and inference, leading to degraded resolution, biased estimates, or reduced robustness of machine-learning-based decision systems \cite{s24102958}.

Modern detectors~\cite{Golling:941076} operate over long time scales and under intense radiation exposure~\cite{butz_2009_57pe1-n8y94,Mudholkar_2019,Aad:2700249}, and their response must be continuously monitored and corrected to preserve the fidelity of physics measurements \cite{Aad:2700249,Mudholkar_2019,Lampen_2011}.
Calibration \cite{Verducci:1177370,Mudholkar_2019,Lampen_2011} and alignment \cite{Ovcharova:1418845,Verducci:1177370} procedures are therefore a core part of detector operations. However, comprehensive recalibration may be resource-intensive and is not always feasible at the frequency that changing detector conditions would ideally require \cite{Laycock_2017,Lampen_2011,B_Pinto_2008}. A key difficulty is that the parameters governing detector degradation are often not directly observable. In many realistic scenarios, one has access to data recorded under nominal (reference) conditions and data recorded under changed conditions, but lacks event-level correspondences or labeled degradation targets \cite{JMLR:v17:15-239,8237506}. This motivates approaches that infer detector-change parameters by exploiting distributional differences between datasets rather than relying on supervised targets or explicit likelihood models \cite{Cranmer2020Frontier,louppe2019adversarial}.

From a machine learning perspective, such non-stationarities correspond to distribution shift \cite{quinonero-candela2008dataset}, also referred to as dataset shift or domain shift, where the data-generating distribution differs between two conditions. While drift monitoring is typically formalized as a decision problem (i.e., whether and when drift occurs), detector systems require estimating the underlying, physically meaningful transformation parameters responsible for the shift (e.g., alignment offsets, attenuation coefficients, or calibration constants) \cite{10.3389/frai.2024.1330257,Ovcharova:1418845,Verducci:1177370}. In detector systems, the objective is not merely to flag a change but to infer interpretable parameters that can be used to correct the data \cite{Verducci:1177370,Laycock_2017,Mudholkar_2019}.

In this work, we propose a novel interpretable calibration approach to tackle sensor degradation as a distribution-matching problem \cite{ganin2016domain}. We introduce a Wasserstein-inspired adversarial parameter-search framework \cite{arjovsky2017wasserstein} for unsupervised recovery of interpretable detector degradation parameters from distribution shift, and demonstrate it on a tracker misalignment toy model and a calorimeter aging task.\footnote{The implementation used in this work is publicly available at \url{https://github.com/tharaaali/wasserstein-parameter-recovery}.} We thus position detector calibration and alignment within a broader class of distribution-shift and distribution-alignment problems. The proposed method is formulated and tested in an offline setting, however, it can in principle be adapted to online calibration scenarios.

\section{Background}
\label{sec:background}

\subsection{HEP detector systems: tracking alignment and calorimeter aging}
\label{subsec:detector_alignment_calibration}

Modern high energy physics (HEP) experiments rely on large scale, high-granularity detector systems designed to reconstruct particle trajectories and measure their energies with high precision. These detectors are typically composed of multiple sub systems, including tracking detectors and calorimeters, each optimized for specific measurement tasks.

\paragraph{Tracking Detectors and Alignment}

Tracking detectors infer charged-particle trajectories from spatial hit measurements recorded in many individual sensor elements. Precise reconstruction requires that the relative positions and orientations of these elements are known in a global reference frame: even when the intrinsic hit resolution is at the level of tens of microns, construction and assembly uncertainties can be much larger and can bias fitted track parameters if uncorrected \cite{Ovcharova:1418845,CERN-LHCC-97-016}. In this context, misalignment is the difference between the true installed detector geometry and the assumed geometry used in reconstruction, and it can systematically bias reconstructed track parameters through biased hit positions in the fit \cite{Airapetian:391176}.

\paragraph{Calorimeters, Aging, and Calibration}

While tracking detectors measure charged-particle trajectories, calorimeters measure particle energy through total absorption \cite{Fabjan2003,Korpachov2018,LHCb2008,ALICE2008,ATLAS2008}. A calorimeter is segmented into cells, each producing a signal proportional to deposited energy. Accurate energy reconstruction requires uniform response across all cells. In an ideal detector, identical particles impinging on different cells should produce identical signals at large statistics. 
In practice, several effects break this uniformity \cite{Fabjan2003,Korpachov2018}. Manufacturing tolerances, electronic variations, temperature fluctuations, and radiation damage introduce cell to cell response variations. Over time, radiation exposure leads to material degradation and signal attenuation—commonly referred to as detector aging. 
Calibration \cite{Fabjan2003,Korpachov2018} procedures aim to restore uniformity by determining correction coefficients for each channel. Calibration is typically established using a combination of hardware procedures (e.g., electronic pulse injection, laser or radioactive sources), test beam measurements, and in-situ physics constraints from well understood processes \cite{Fabjan2003,Korpachov2018}.
Over time, calorimeter response can drift due to aging and radiation damage in active materials and readout components, altering cell wise gains or attenuation and necessitating periodic recalibration to maintain uniformity and energy scale stability \cite{Fabjan2003}.
Because the detector comprises many channels, maintaining calibration becomes a large scale parameter estimation problem, where physically meaningful parameters (e.g., per-cell response coefficients) must be updated from data under realistic constraints on dedicated calibration time and resources \cite{Fabjan2003,Korpachov2018}.

\subsection{Applications of Generative Machine Learning Models in High Energy Physics}
\label{subsec:gan_hep}

Deep generative models have become a central component of modern machine-learning toolkits for HEP because they can learn high-dimensional detector- and event-level distributions and provide efficient sampling procedures~\cite{HASHEMI2024100092}. A major driver is fast detector simulation: surrogate generative models can approximate the output of computationally
expensive Monte Carlo pipelines while retaining statistical features needed for downstream analyses~\cite{HASHEMI2024100092,deOliveira2017}. At the same time, high fidelity deployment requires careful validation and uncertainty treatment, since mismodeling can bias physics conclusions even when one dimensional projections look reasonable \cite{10.21468/SciPostPhys.10.6.139}.

Recent reviews organize state of the art surrogate models by architecture~\cite{kingma2022autoencodingvariationalbayes} class, including GAN based models~\cite{GAN1}, variational autoencoders (VAEs), normalizing flows~\cite{9089305}, diffusion/score-based models~\cite{NEURIPS2020_4c5bcfec}, and their hybrids, as well as by the data representation and conditioning strategy used in different detectors and benchmarks~\cite{HASHEMI2024100092}. In practice, GANs have been widely explored for calorimeter and detector-response emulation, with substantial gains in generation speed compared to traditional simulation chains~\cite{chang2021dual,RATNIKOV2023167591,Anderlini:2024LamarrEPJWebConf,matchev2022uncertainties,deOliveira2017,chang2021quantum}. However, their use also raises questions of statistical fidelity and uncertainty propagation, motivating rigorous validation protocols~\cite{10.21468/SciPostPhys.10.6.139}.

Generative models are also used as components of analysis workflows, especially in weakly supervised and unsupervised settings. Autoencoder-based approaches, including conditional VAEs~\cite{pol2019anomaly}, have been studied for anomaly detection at the LHC \cite{lavizzari2023variational}, where deviations from the learned manifold may indicate rare or unexpected processes. More broadly, generative modeling can support model-independent searches and background modeling under limited supervision~\cite{HASHEMI2024100092}.

Beyond event generation and simulation acceleration, GANs have also been explored as tools for detector calibration and parameter tuning. In this context, the adversarial objective is used not to generate new samples, but to enforce consistency between observed data distributions and reference expectations \cite{stoye2018likelihood,ramazyan2024global}. Related ideas have been investigated using likelihood-based methods and supervised neural networks for parameter estimation, including generator tuning and detector-response modeling~\cite{febrianti2021parameter,andreassen2021parameter}. Track-based alignment provides a classical example of large-scale parameter inference in HEP, where geometric constants are extracted from reconstructed tracks through iterative optimization and must address scalability and parameter degeneracies~\cite{ATLAS-CONF-2011-012,Bruckman:835270,Bocci:1039585}.

This motivates distribution level strategies \cite{ZUO202237} for parameter inference that do not require event level correspondences. Adversarial objectives, and in particular Wasserstein-based critics, provide a practical way to quantify and minimize mismatch between high dimensional distributions~\cite{pmlr-v70-arjovsky17a,WGAN-GP}. In this work, we adopt this perspective by using a deterministic transformation with trainable weights corresponding to calibration or alignment parameters, optimized using a distribution-matching objective.

\subsection{Generative Adversarial Networks}
\label{gan_intro}

A GAN is composed of two components trained simultaneously: a generator \(G\), which produces samples intended to resemble data drawn from a target distribution, and a discriminator \(D\), which evaluates whether a given sample originates from the true data distribution or from the generator. Rather than relying on an explicit likelihood, GANs learn by enforcing statistical indistinguishability between generated and real samples~\cite{GAN1}.

During training, the generator maps latent variables \(z\), sampled from a predefined distribution \(p_z(z)\), to synthetic samples \(G(z)\). The discriminator receives either real samples \(x \sim p_{\text{data}}(x)\) or generated samples \(G(z)\) and outputs a scalar score reflecting its confidence in the sample’s authenticity. The two networks are optimized in opposition: the discriminator improves its ability to separate real and generated data, while the generator adapts to produce samples that are increasingly difficult to distinguish.

This adversarial interaction is commonly formulated as a minimax optimization problem~\cite{GAN1}:
\begin{equation}
\min_G \max_D 
\mathbb{E}_{x \sim p_{\text{data}}(x)}[\log D(x)] +
\mathbb{E}_{z \sim p_z(z)}[\log (1 - D(G(z)))] ,
\end{equation}
where the equilibrium corresponds to the generator reproducing the target data distribution. Although this formulation has proven effective in many applications, it is known to suffer from optimization instabilities and sensitivity to model design and hyperparameters, particularly when the real and generated distributions have limited overlap~\cite{arjovsky2017wasserstein}.

The Wasserstein GAN (WGAN) is proposed to address the optimization challenges inherent to the original GAN formulation by replacing the discriminator-based divergence with a distance metric grounded in optimal transport theory~\cite{arjovsky2017wasserstein}. Instead of estimating a probability of authenticity, the discriminator is reformulated as a \emph{critic} \(C\), which assigns real-valued scores to samples. The difference between expected critic scores for real and generated data provides an estimate of the Wasserstein-1 (Earth Mover’s) distance between their distributions.

The WGAN objective is given by

\begin{equation}
\min_G \max_{C \in \mathcal{C}} \mathbb{E}_{x \sim p_{\text{data}}(x)}[C(x)] - \mathbb{E}_{z \sim p_z(z)}[C(G(z))],
\end{equation}
where \(\mathcal{C}\) denotes the set of 1-Lipschitz functions. Enforcing this constraint is essential for the validity of the Wasserstein distance and is commonly achieved through techniques such as weight clipping~\cite{arjovsky2017wasserstein}, gradient penalty regularization~\cite{WGAN-GP}, or spectral normalization~\cite{SN-GAN}.

By providing more informative gradients and a loss function that correlates with distributional mismatch, WGAN improves training stability and robustness compared to the original GAN framework~\cite{arjovsky2017wasserstein}. Importantly, the critic output offers a meaningful quantitative measure of how far two distributions are from one another, making WGAN a natural choice for problems centered on distribution alignment rather than explicit sample generation.

\subsection{Alignment and calibration as distribution matching parameter inference}
\label{wgan_parameters}

Tracking alignment and calorimeter calibration share the same mathematical structure: unknown physical parameters (geometric shifts or response coefficients) must be inferred from data, while only indirect observables (hit positions or energy deposits) are accessible. In tracking, misalignment alters the spatial distribution of residuals and reconstructed track parameters~\cite{Bruckman:835270,ATLAS-CONF-2011-012}. In calorimetry, imperfect equalization and long term detector aging modify the statistical distribution of energy deposits and shower shapes, motivating periodic calibration and monitoring to preserve the stability of reconstructed energies~\cite{Fabjan2003}.

In both cases, detector deformations induce a distributional shift between a nominal (reference) dataset and a distorted (changed/damaged) dataset. Traditional procedures often rely on explicit residual-based objectives~\cite{ATLAS-CONF-2011-012,ATLAS-CONF-2012-141}, or on external reference signals and well controlled physics samples for calorimeter calibration~\cite{Fabjan2003}. This motivates complementary approaches that operate directly at the level of data distributions: instead of minimizing residuals event-by-event, one may infer the underlying physical parameters by matching the full high-dimensional distribution of observables between a reference and a shifted dataset.

While Wasserstein GANs (WGANs) are most often employed for synthetic data generation, their formulation as a distribution matching framework enables alternative modes of use beyond sample synthesis~\cite{arjovsky2017wasserstein}. In particular, the generator can be used as a deterministic transformation whose trainable weights represent transformation parameters applied to observed data~\cite{8237506,JMLR:v17:15-239}. In this setting, the adversarial objective enforces agreement between transformed and reference distributions, and the learned parameters implicitly capture the systematic effects responsible for their mismatch. This perspective is especially relevant for detector studies, where effects such as response changes due to radiation exposure and aging~\cite{Fabjan2003}, as well as geometric misalignment~\cite{Bruckman:835270,ATLAS-CONF-2011-012}, manifest as structured changes in measured data distributions~\cite{andreassen2021parameter,febrianti2021parameter}. By minimizing the Wasserstein distance between transformed and nominal detector responses, the WGAN framework enables unsupervised inference of physically interpretable parameters without requiring labeled data or explicit likelihood models.

In the present work, we adopt this paradigm by embedding detector-related parameters within the generator and using the critic exclusively as a distributional consistency measure. This shifts the role of adversarial learning from data synthesis to parameter recovery, providing a flexible and scalable approach for data-driven detector calibration and alignment studies.

\section{Method and architecture}
\label{sec:method}

We propose a data-driven framework for inferring physically interpretable parameters that characterize systematic changes between two data taking regimes. The method is designed for scenarios in which a reference dataset, corresponding to an undamaged or nominal detector state, is available alongside a second dataset exhibiting distortions induced by aging, misalignment, or other systematic effects. Rather than explicitly modeling these effects through parametric likelihoods, we formulate the problem as one of distribution alignment and parameter recovery. To achieve this aim, we adopt the adversarial training principle but reinterpret the role of the generator: instead of producing new data samples, it consists of a parameterized transformation applied directly to existing data, the transformation parameters are treated as trainable model variables. From a conventional perspective, the task of the generator is to correct the damaged detector responses. Specifically, we consider two distributions: a reference distribution \(p_{\mathrm{ref}}\), representing undamaged or nominal detector responses, and a target distribution \(p_{\mathrm{chg}}\), corresponding to damaged or systematically altered data. A deterministic transformation \(T_{\boldsymbol{\theta}}(\cdot)\), parameterized by \(\boldsymbol{\theta}\), is applied to samples drawn from one of the distributions. The objective is to adjust \(\boldsymbol{\theta}\) such that the transformed distribution aligns statistically with the opposing distribution. Depending on the chosen formulation, this corresponds to matching \(T_{\boldsymbol{\theta}}(p_{\mathrm{chg}})\) to \(p_{\mathrm{ref}}\), or equivalently matching \(p_{\mathrm{chg}}\) to \(T_{\boldsymbol{\theta}}(p_{\mathrm{ref}})\).

Inspired by the WGAN algorithm, a critic network is employed to quantify the discrepancy between the transformed and target distributions. By minimizing the Wasserstein distance estimate, the model directly optimizes the transformation parameters that best explain the observed distributional shift. This yields a stable optimization procedure and a scalar objective that correlates with distributional mismatch.

\begin{figure*}[t]
\includegraphics[width=1\hsize]{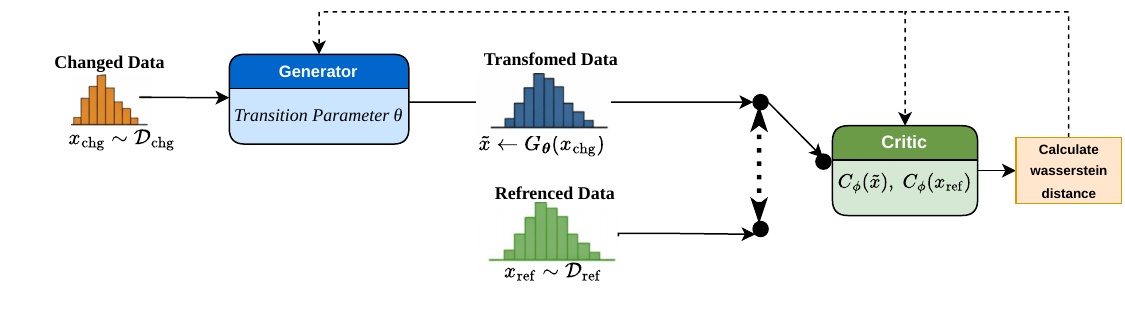}
\caption{Architecture of the proposed WGAN-inspired adversarial parameter-search framework.}
\label{fig:archMod}
\end{figure*}

Fig.~\ref{fig:archMod} illustrates the overall architecture. While implementation details vary between applications, the core components a parameterized transformation, an adversarial critic, and a Wasserstein based objective remain unchanged throughout this work.

\subsection{Operational principle}
\label{subsec:how_it_works}
The method operates by iteratively updating the transformation parameters \(\boldsymbol{\theta}\) to minimize the discrepancy between transformed and reference data distributions. At each training step, minibatches of samples are drawn from the reference and changed datasets. The transformation \(T_{\boldsymbol{\theta}}\) is applied to a set of samples, and the critic evaluates both transformed and untransformed samples to estimate the Wasserstein distance between their distributions.
The transformation parameters are optimized to minimize this distance, thereby driving the transformed distribution toward statistical consistency with the reference. Importantly, no direct supervision on \(\boldsymbol{\theta}\) is required, and no explicit correspondence between individual samples in the two datasets is assumed. The inferred parameters are learned from population-level distributional agreement.
This adversarial parameter-search framework is symmetric by construction. Depending on the direction of the applied transformation, it can be used either to recover undamaged detector responses from altered data or to impose controlled distortions on nominal data. Consequently, the same architecture supports both calibration-oriented inference and controlled simulation studies within a unified framework.
\subsection{Parameterized transformation model}
\label{subsec:theta_definition}
Let $\mathbf{y}\in\mathbb{R}^d$ denote a generic detector observation vector, with $\mathbf{y}\sim p_{\mathrm{chg}}$ for changed data and $\mathbf{y}\sim p_{\mathrm{ref}}$ for reference data. We introduce a deterministic transformation
\begin{equation}
T_{\boldsymbol{\theta}}:\mathbb{R}^d\rightarrow\mathbb{R}^d,
\end{equation}
parameterized by $\boldsymbol{\theta}$, where $\boldsymbol{\theta}$ has a direct physical interpretation.
\paragraph{Tracker toy model}
In the tracker toy example, we consider three measurement planes and use a single spatial coordinate. Let $\mathbf{h}=(h_1,h_2,h_3)\in\mathbb{R}^3$ denote the hit coordinates on planes $\ell=1,2,3$. A misalignment is injected by shifting plane $\ell=2$ by an unknown $\Delta x_2$. The correction transformation applies an opposite shift to the second component:
\begin{equation}
T_{\Delta x_2}(\mathbf{h}) = (h_1,\ h_2 + \Delta x_2,\ h_3).
\end{equation}
\paragraph{Calorimeter aging}
Let $\mathbf{e}\in\mathbb{R}^N$ denote the vector of cell-wise energy deposits for an event in the changed dataset, with $e_i$ the deposit in cell $i$. Aging is modeled as a cell-wise attenuation with coefficients $A_i\in(0,1]$ such that $e^{(i)}_{\mathrm{aged}} = A_i\, e^{(i)}_{\mathrm{true}}$. The calibration transformation rescales cell energies to undo the attenuation:
\begin{equation}
T_{\boldsymbol{\alpha}}(\mathbf{e})_i = \frac{e_i}{\alpha_i},
\qquad
\boldsymbol{\alpha}=(\alpha_1,\ldots,\alpha_N),\ \alpha_i\in(0,1].
\end{equation}
Ideally, $\alpha_i$ converges to the true aging coefficient $A_i$.
While these two cases differ in dimensionality and physical interpretation, both can be treated uniformly within the proposed framework by treating \(\boldsymbol{\theta}\) as trainable parameters of a deterministic transformation module \(T_{\boldsymbol{\theta}}\), optimized through adversarial distribution matching.

\subsection{Adversarial objective}
\label{subsec:objective}

We introduce a critic $C_\phi$ parameterized by $\phi$, constrained to be 1-Lipschitz. The WGAN objective for aligning corrected changed data with the reference data is
\begin{equation}
\min_{\boldsymbol{\theta}}
\max_{\phi\in\mathcal{C}}
\;
\mathbb{E}_{\mathbf{y}\sim p_{\mathrm{ref}}}\!\left[C_\phi(\mathbf{y})\right]
-
\mathbb{E}_{\mathbf{y}\sim p_{\mathrm{chg}}}\!\left[
C_\phi\!\left(T_{\boldsymbol{\theta}}(\mathbf{y})\right)
\right],
\end{equation}
where $\mathcal{C}$ denotes the set of 1-Lipschitz functions. The critic is trained to maximize the objective, while $\boldsymbol{\theta}$ is trained to minimize it, driving $T_{\boldsymbol{\theta}}(p_{\mathrm{chg}})$ toward
$p_{\mathrm{ref}}$.

In both cases, the critic is trained to maximize the objective (increase the estimated Wasserstein distance), while $\boldsymbol{\theta}$ is trained to minimize it (reduce the distance by improving the transformation).
 
\begin{algorithm}[H]
\caption{Adversarial parameter search}
\label{alg:adv_param_search}
\begin{algorithmic}[1]
\REQUIRE Reference dataset $\mathcal{D}_{\mathrm{ref}}$,
         changed dataset $\mathcal{D}_{\mathrm{chg}}$,
         critic $C_\phi$,
         transformation $T_{\boldsymbol{\theta}}$,
         clipping parameter $c$,
         number of critic updates $n_{\mathrm{critic}}$
\ENSURE Learned parameters $\boldsymbol{\theta}$

\STATE Initialize critic parameters $\phi$
\STATE Initialize transformation parameters $\boldsymbol{\theta}$
      (e.g., $\alpha_i \leftarrow 1$ for calorimeter; $\Delta x_2 \leftarrow 0$ for tracker)

\FOR{each training iteration}
    \FOR{$n_{\mathrm{critic}}$ critic updates}
        \STATE Sample minibatch $x_{\mathrm{ref}} \sim \mathcal{D}_{\mathrm{ref}}$
        \STATE Sample minibatch $x_{\mathrm{chg}} \sim \mathcal{D}_{\mathrm{chg}}$
        \STATE Correct changed samples: $\tilde{x} \leftarrow T_{\boldsymbol{\theta}}(x_{\mathrm{chg}})$
        \STATE Critic loss (to \emph{minimize} by gradient descent):
        \[
        \mathcal{L}_C =
        - \Big(
        \mathbb{E}[C_\phi(x_{\mathrm{ref}})]
        -
        \mathbb{E}[C_\phi(\tilde{x})]
        \Big)
        \]
        \STATE Update $\phi$ by descending $\nabla_\phi \mathcal{L}_C$
        \STATE Clip critic weights: $\phi \leftarrow \mathrm{clip}(\phi,-c,c)$
    \ENDFOR

    \STATE Sample minibatch $x_{\mathrm{chg}} \sim \mathcal{D}_{\mathrm{chg}}$
    \STATE Correct changed samples: $\tilde{x} \leftarrow T_{\boldsymbol{\theta}}(x_{\mathrm{chg}})$
    \STATE Generator/transformation loss:
    \[
    \mathcal{L}_T = -\,\mathbb{E}[C_\phi(\tilde{x})]
    \]
    \STATE Update $\boldsymbol{\theta}$ by descending $\nabla_{\boldsymbol{\theta}} \mathcal{L}_T$
\ENDFOR

\RETURN $\boldsymbol{\theta}$
\end{algorithmic}
\end{algorithm}

\section{Experiments}
\label{sec:experiments}
\subsection{Case study I: Tracker Toy Model}
\label{subsec:tracker_toy}
As a first validation of the proposed adversarial parameter-search framework, we consider a simplified tracking-detector toy model. The purpose of this study is not to provide a realistic alignment solution, but rather to demonstrate that the method can reliably recover a known geometric misalignment parameter in a controlled and interpretable setting.

\begin{figure}
    \centering
\includegraphics[width=\linewidth]{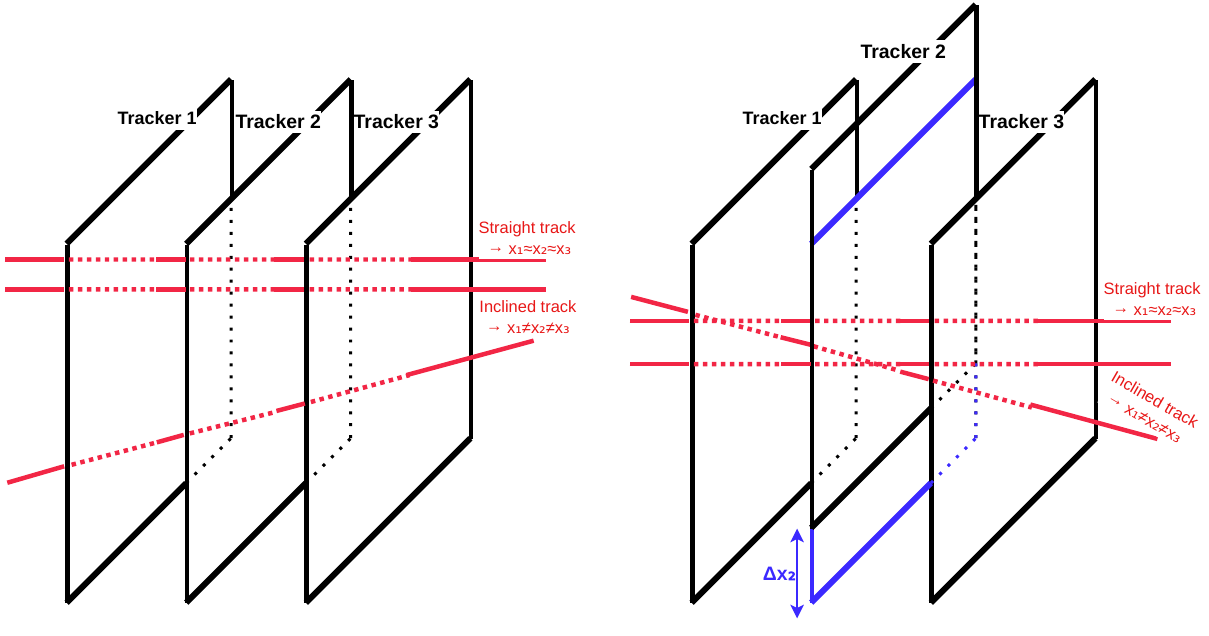}
    \caption{Tracker toy model, a misalignment is injected by shifting the second plane by $\Delta x_2$, producing a systematic change in hit patterns (and residual distributions) relative to the nominal geometry.}
    \label{fig:tracker_sim}
\end{figure}
The toy detector consists of three tracking layers (Fig.~\ref{fig:tracker_sim}) that record one-dimensional hit coordinates. Two types of trajectories are simulated: straight tracks and inclined tracks. In the reference configuration, all three planes are perfectly aligned, yielding consistent geometric relations across layers.
Two datasets are constructed:
\begin{itemize}
    \item \textbf{Reference dataset} ($\mathcal{D}_{\mathrm{ref}}$): generated with all tracker layers at their nominal positions.

    \item \textbf{Changed dataset} ($\mathcal{D}_{\mathrm{chg}}$): generated by shifting the second tracking plane by a constant offset $\Delta x_2$ while keeping the remaining planes fixed.
\end{itemize}
\paragraph{Generator as a single parameter offset model}
In this setup, we have a reference (undamaged) distribution \(p_{\mathrm{ref}}\) and a changed/distorted distribution \(p_{\mathrm{chg}}\). Samples from the changed dataset are passed through the generator, which in this case is as a deterministic correction transform \(T_{\boldsymbol{\theta}}\) with a single trainable scalar parameter \(\boldsymbol{\theta}=\widehat{\Delta x_2}\). Each sample is a hit tuple
\(\mathbf{h}=(h_1,h_2,h_3)\in\mathbb{R}^3\), where \(h_\ell\) is the measured
\(x\) coordinate on plane \(\ell\in\{1,2,3\}\). The transformation acts on the second coordinate and outputs a corrected sample \(\tilde{\mathbf{h}}=T_{\widehat{\Delta x_2}}(\mathbf{h})\):
\begin{equation}
\tilde{\mathbf{h}} = T_{\widehat{\Delta x_2}}(\mathbf{h})
= (h_1,\ h_2 + \widehat{\Delta x_2},\ h_3).
\end{equation} Training therefore directly yields \(\widehat{\Delta x_2}\) as the inferred misalignment correction. The critic \(C_\phi\) is a neural network trained with the WGAN objective to maximize the separation between reference samples \(\mathbf{h}_{\mathrm{ref}}\sim p_{\mathrm{ref}}\) and corrected samples \(\tilde{\mathbf{h}}\sim T_{\widehat{\Delta x_2}}(p_{\mathrm{chg}})\), providing an estimate of the Wasserstein distance. The parameter \(\widehat{\Delta x_2}\) is updated in the opposite direction to minimize this estimated distance, thereby aligning \(T_{\widehat{\Delta x_2}}(p_{\mathrm{chg}})\) with \(p_{\mathrm{ref}}\). At convergence, \(\widehat{\Delta x_2}\) approaches $-\Delta x_2$, since this value best aligns the corrected and reference hit distributions.
To study the sensitivity to a systematic distortion, a fixed shift value of $\Delta x_2 = 0.01$ is applied to the detector coordinates. Each dataset contains only hit coordinates and does not include labels or track level correspondences between the reference and changed samples, ensuring a fully unsupervised setup based exclusively on distribution-level information.
To assess robustness against measurement uncertainty, independent Gaussian noise is added to the hit coordinates in the changed dataset. The noise standard deviation $\sigma$ is varied over several orders of magnitude, ranging from negligible detector noise to values comparable to the applied shift.

\subsection{Case study II: Calorimeter Aging}
\label{subsec:calorimeter}

This case study investigates whether the proposed adversarial parameter search framework can recover cell wise aging effects in a fully unsupervised manner by exploiting distributional differences between undamaged and aged calorimeter responses.

\begin{figure}[t]
    \centering
    \includegraphics[width=0.8\linewidth]{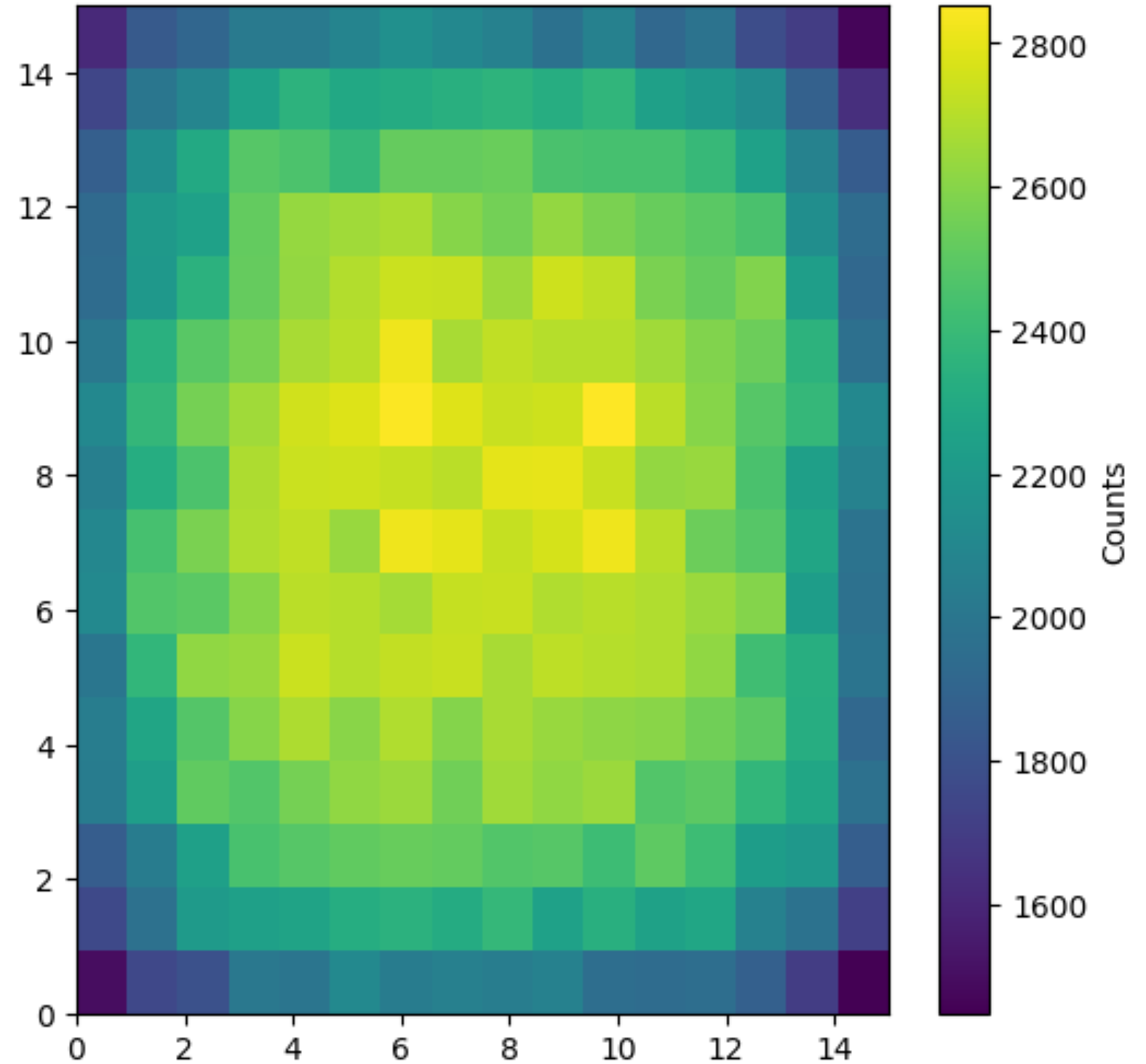}
    \caption{Hit activity map in the central $16\times16$ cell region (counts summed over events).}
    \label{fig:calo_hitmap}
\end{figure}

\paragraph{Simulation and data representation}
Monte Carlo simulations are performed using the \textsc{Geant4} toolkit~\cite{GEANT4:2002zbu}. We generate $10^5$ events of single $\pi^{-}$ mesons with an exponential kinetic-energy spectrum and a flat transverse impact distribution; the particle momentum is oriented perpendicular to the upstream face of the calorimeter. The detector geometry corresponds to a highly granular sampling calorimeter with alternating 20\,mm lead absorber plates and 5\,mm polystyrene scintillator tiles, segmented into a transverse cell grid. To reduce edge effects and shower leakage, we restrict the analysis to the central $16\times16$ cells in each sensitive layer. Fig.~\ref{fig:calo_hitmap} shows the corresponding transverse hit-activity map aggregated over the dataset.

Each event is represented as a 3D tensor
\[
X \in \mathbb{R}^{L\times H\times W},
\]
where $L$ is the number of sensitive layers and $(H,W)$ correspond to the selected transverse cell grid; $X_{\ell,u,v}$ is the energy deposit in cell $(\ell,u,v)$. We denote by $p_{\mathrm{ref}}$ and $p_{\mathrm{chg}}$ the distributions of undamaged and aged events, respectively.
\paragraph{Aging model and datasets}
Two complementary datasets are constructed:
\begin{itemize}
  \item \textbf{Undamaged dataset} ($\mathcal{D}_{\mathrm{ref}}$): events   $X \sim p_{\mathrm{ref}}$ simulated without aging effects.
  \item \textbf{Changed dataset} ($\mathcal{D}_{\mathrm{chg}}$): events   $X_{\mathrm{chg}} \sim p_{\mathrm{chg}}$ obtained by applying cell-wise signal attenuation to emulate radiation damage.
\end{itemize}

\begin{figure}[t]
            \centering
            \includegraphics[width=\linewidth]{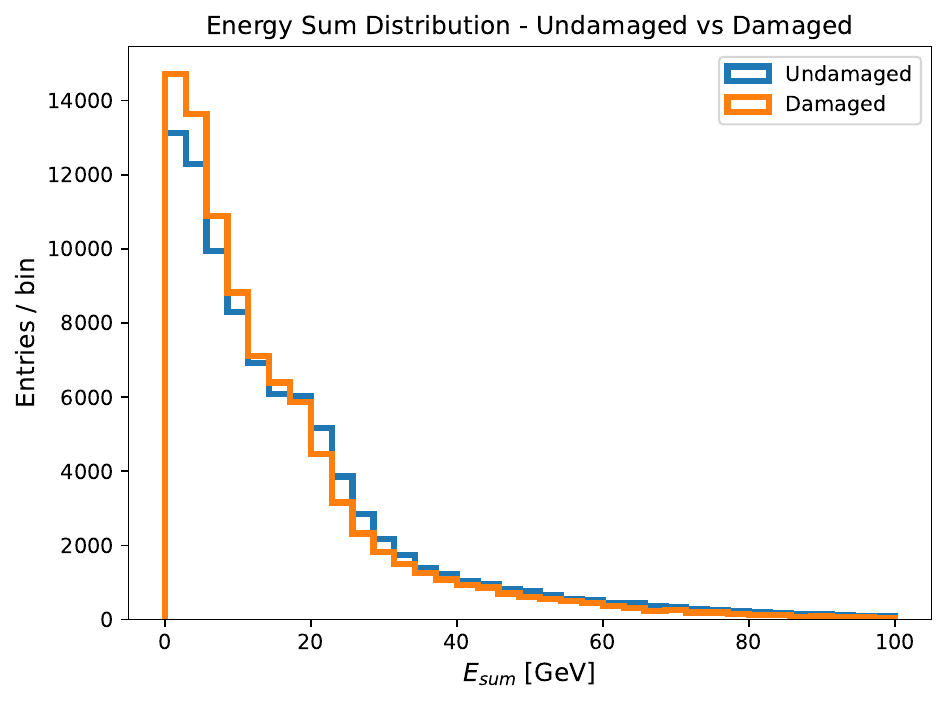}
            \caption{Distribution of total deposited energy per event for undamaged and aged calorimeter responses.}
            \label{fig:EnergySumDistribution}
        \end{figure}

We model aging through a cell-wise coefficient field \[
\mathbf{A} \in (0,1]^{L\times H\times W},
\]
so that an undamaged event $X$ is mapped to an aged event by element-wise multiplication:
\begin{equation}
X_{\mathrm{chg}} = T_{\mathbf{A}}^{\mathrm{age}}(X) \equiv \mathbf{A}\odot X,
\qquad (X_{\mathrm{chg}})_{\ell,u,v}=A_{\ell,u,v}\,X_{\ell,u,v},
\end{equation}
where $\odot$ denotes the Hadamard product. In our simulation setup, the coefficients are generated as
\begin{equation}
A_i = f(n_i) + \mathcal{N}(0,\epsilon),
\end{equation}
where $i$ indexes calorimeter cells, $n_i$ is the total hit count accumulated by cell $i$ over the dataset, $f(\cdot)$ is a monotone decay model assigning smaller coefficients to more frequently activated cells, and $\epsilon$ controls cell-to-cell miscalibration noise (unless stated otherwise, $\epsilon=0.01$).
Fig.~\ref{fig:EnergySumDistribution} illustrates the effect of aging by comparing the per event total deposited energy, $E_{\mathrm{sum}}=\sum_i E^{(i)}$, for undamaged and aged samples, where a noticeable shift is observed.

\paragraph{Generator as a calibration field}
The generator is the parameterized transformation $T_{\boldsymbol{\theta}}$ whose trainable parameters are the calibration coefficients themselves:
\[
\boldsymbol{\theta}\equiv\widehat{\mathbf{A}}\in (0,1]^{L\times H\times W}.
\]
Operationally, $\widehat{\mathbf{A}}$ is implemented as a learnable tensor matching the calorimeter granularity (a 3D parameter field). Given an aged event $X_{\mathrm{chg}}\sim p_{\mathrm{chg}}$, the generator applies an element wise correction and outputs a calibrated event
\begin{equation}
\widetilde{X} = T_{\widehat{\mathbf{A}}}(X_{\mathrm{chg}})
\equiv X_{\mathrm{chg}} \oslash \widehat{\mathbf{A}},
\qquad \widetilde{X}_{\ell,u,v}=\frac{(X_{\mathrm{chg}})_{\ell,u,v}}{\widehat{A}_{\ell,u,v}},
\label{eq:calo_correction}
\end{equation}
where $\oslash$ denotes element wise division. 
To ensure numerical stability, we constrain $\widehat{A}_{\ell,u,v}\in[\alpha_{\min},1]$ via clamping after each update.

The critic network $C_\phi$ is trained with the WGAN objective to estimate the Wasserstein distance between the corrected changed distribution $T_{\widehat{\mathbf{A}}}(p_{\mathrm{chg}})$ and the reference distribution $p_{\mathrm{ref}}$. Training alternates between updating the critic to maximize the estimated Wasserstein discrepancy and updating the generator parameters $\widehat{\mathbf{A}}$ to minimize it. At convergence,
$\widehat{\mathbf{A}}$ provides an estimate of the underlying aging field and thus yields cell-wise calibration coefficients directly as learned generator parameters.

Although aging primarily affects the global energy scale, global energy-sum projections alone are insufficient for identifying cell wise effects. Instead, the information that constrains $\widehat{\mathbf{A}}$ is encoded in the high-dimensional structure of cell level depositions and their correlations, which are matched implicitly by the critic during adversarial training. The overall architecture of the WGAN-inspired calibration model is shown in Fig.~\ref{fig:archMod}.

\section{Results and discussion}
\label{sec:calo_results}

\begin{figure}[t]
            \centering
            \includegraphics[width=0.8\linewidth]{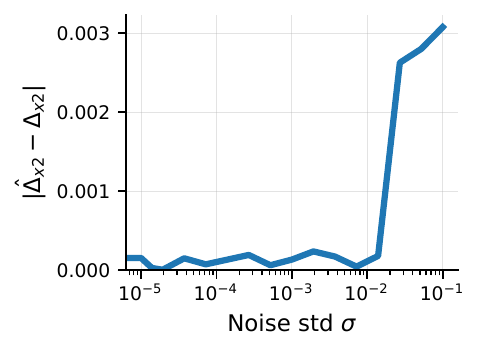}
            \caption{Absolute error of the estimated tracker shift $|\widehat{\Delta x} - \Delta x|$ as a function of the Gaussian noise standard deviation $\sigma$. Accurate recovery is observed for low and moderate noise levels, while the estimation error increases rapidly once the noise becomes comparable to the applied shift.}
            \label{fig:tracker_shift_010}
        \end{figure}

\paragraph{Tracker toy model}
Fig.~\ref{fig:tracker_shift_010} reports the absolute estimation error $|\widehat{\Delta x}-\Delta x|$ as a function of the Gaussian noise standard deviation $\sigma$ for injected misalignment magnitudes. For low and moderate noise levels, the inferred shift remains close to the true value, with errors below $10^{-3}$. As $\sigma$ approaches the scale of the injected shift, the error increases rapidly, consistent with the loss of identifiable geometric information in the hit tuples. Overall, the toy study demonstrates that the proposed distribution matching objective can recover a simple, physically interpretable alignment parameter when the distortion is not dominated by measurement noise.

\paragraph{Calorimeter aging}
We first examine the effect of simulated aging on the calorimeter response. Fig.~\ref{fig:EnergySumDistribution} compares the total reconstructed energy ($E_{\mathrm{sum}}$) distributions for the undamaged and damaged datasets. The damaged distribution exhibits an overall suppression of the energy scale, as expected from cell wise attenuation. At the same time, the distribution shape changes modestly, indicating that a one dimensional global observable alone provides limited sensitivity to spatially varying aging and that additional information must be exploited from the multi cell response patterns.

\begin{figure}[t]
    \includegraphics[width=\hsize]{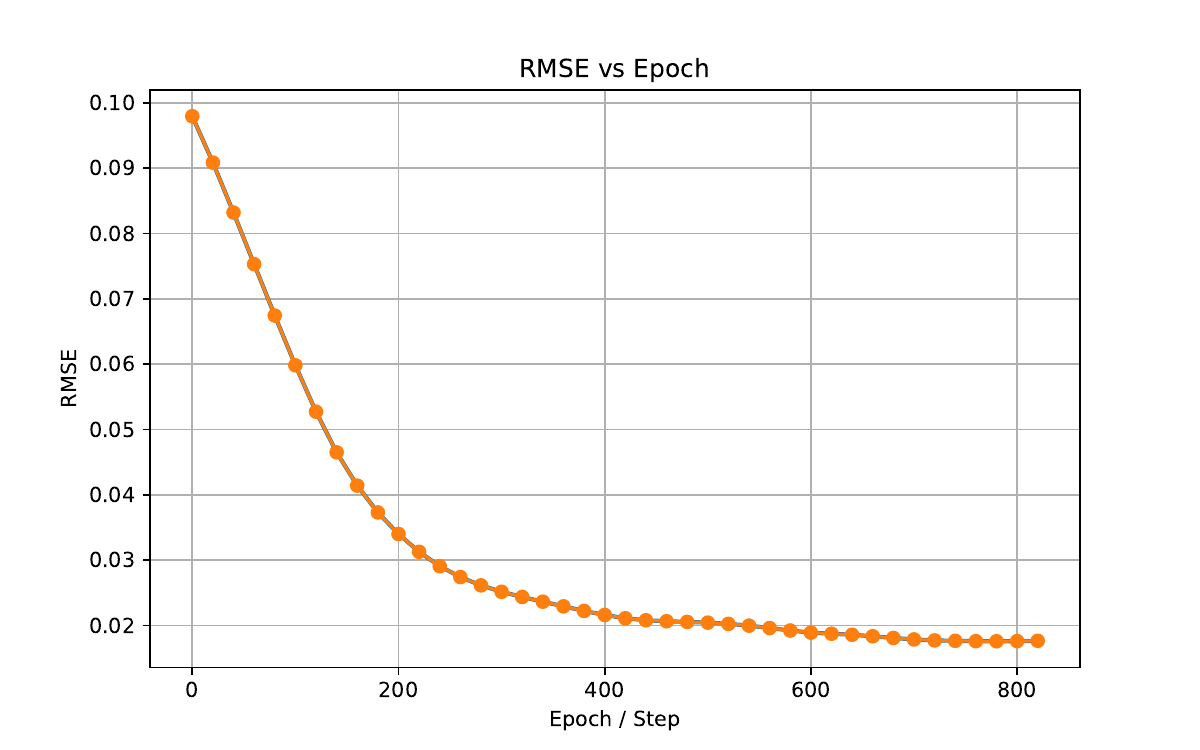}
    \caption{RMSE between inferred and injected aging coefficients.}
    \label{fig:rmse_curve}
\end{figure}

The training dynamics of the WGAN-inspired calibration model are summarized in Fig.~\ref{fig:rmse_curve}, which shows the RMSE between the predicted and injected aging coefficients as a function of epoch. The error decreases during training and reaches a stable plateau after roughly $650$-$700$ epochs, indicating consistent convergence under the chosen optimization setup.

\begin{figure}[t]
    \includegraphics[width=\hsize]{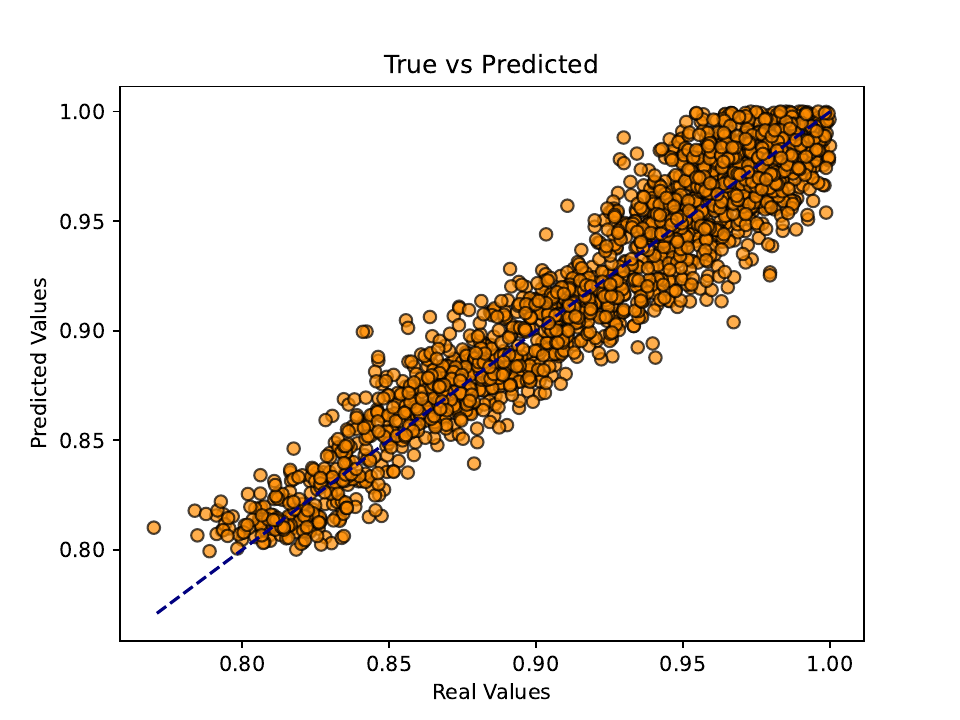}
    \caption{Cell wise comparison of true and inferred aging coefficients.}
    \label{fig:TrueVsPred}
\end{figure}

\begin{figure}[t]
    \centering
    \includegraphics[width=\hsize]{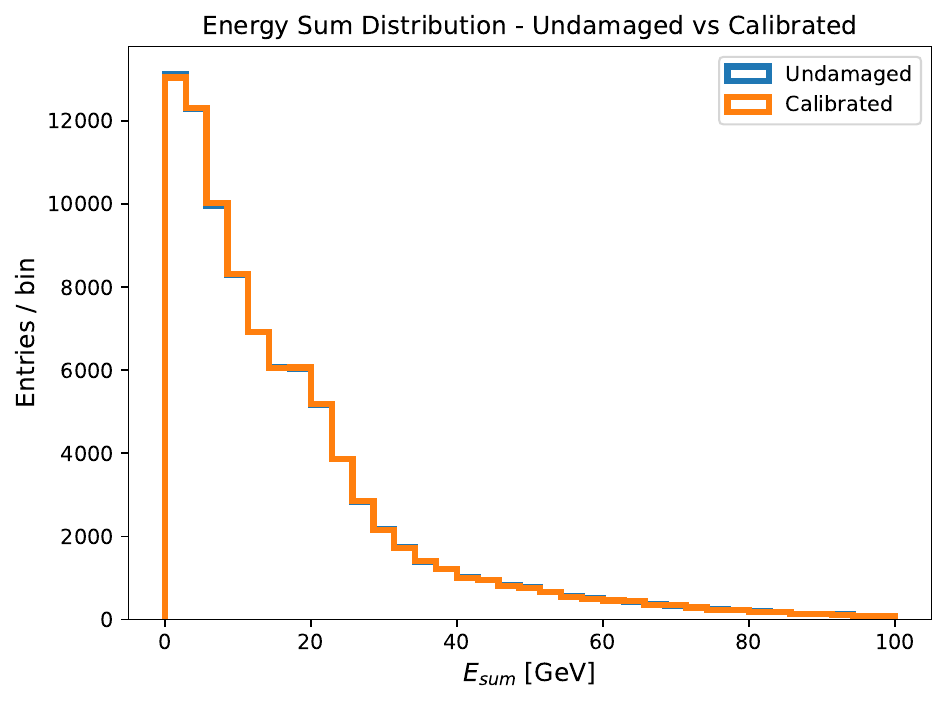}
    \caption{Energy-sum distributions for undamaged and calibrated calorimeter responses.}
    \label{fig:EnergySumCalibrated}
\end{figure}

Fig.~\ref{fig:TrueVsPred} compares the inferred coefficients to the injected values on a cell by cell basis. The estimates follow a clear positive correlation, with most cells clustered near the diagonal, showing that the model recovers the dominant aging trend from distribution-level constraints. The largest deviations are observed for the most strongly attenuated cells, where reduced signal amplitudes and the injected stochastic perturbations limit the information available for parameter identification. We do not interpret these deviations as a failure of calibration, but rather as an expected limitation of inferring fine grained, per cell fluctuations in an unsupervised setting with finite statistics and noisy coefficients. The proposed method achieves a mean reconstruction error of $\mathrm{RMSE}=0.0176 \pm 0.0005$ and a mean coefficient of determination of $R^2 = 0.896 \pm 0.007$. 
A simple per cell baseline that estimates aging as the ratio of mean responses, $ \widehat{A}_i^{\mathrm{mean}}=\frac{\langle E_i^{\mathrm{chg}}\rangle}{\langle E_i^{\mathrm{ref}}\rangle}$,
yields a  worse agreement with the injected coefficients of with $\mathrm{RMSE}=0.0562 \pm 0.0008$. This comparison indicates that the proposed distribution level adversarial calibration extracts significantly more information from the full multi cell response than can be captured by matching only per cell mean energies.
To assess the practical effect of the inferred parameters, we apply the learned correction to the damaged dataset and recompute the energy sum distribution. Fig.~\ref{fig:EnergySumCalibrated} shows that the calibrated distribution moves closer to the undamaged reference over the full energy range. This indicates that the recovered coefficients compensate the dominant radiation-induced attenuation at the level of a global summary observable, while not guaranteeing perfect cell by cell recovery of the injected stochastic component.

\begin{figure}[t]
    \centering
    \includegraphics[width=\hsize]{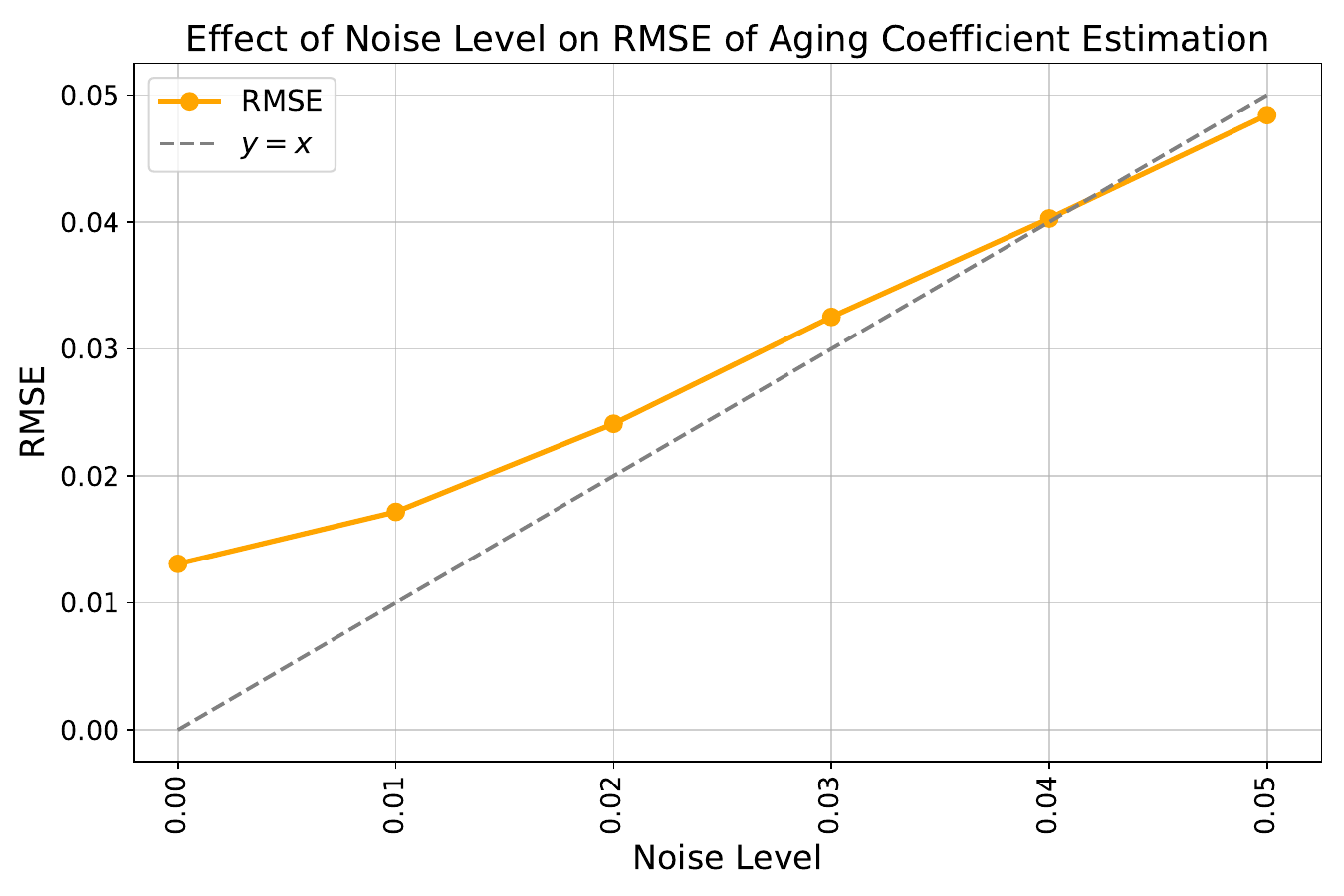}
    \caption{Effect of stochastic coefficient noise $\epsilon$ on the RMSE of aging-coefficient estimation.}
    \label{fig:eps_vs_rmse}
\end{figure}

Finally, we evaluate sensitivity to uncertainty in the aging coefficients by varying the standard deviation $\epsilon$ of the additive Gaussian perturbation in the aging model over $\epsilon\in[0,0.05]$. As shown in Fig.~\ref{fig:eps_vs_rmse}, the RMSE between inferred and injected coefficients increases monotonically with $\epsilon$. This trend is expected: larger cell to cell stochastic variations reduce the identifiability of individual attenuation factors from distribution matching alone, because multiple coefficient configurations can yield similar aggregate shower statistics.

In the deterministic case ($\epsilon=0$), the model attains its lowest error, $\mathrm{RMSE}=0.01573$, which reflects the achievable accuracy for recovering the global aging trend $f(n_i)$ under the given simulation and training setup. As $\epsilon$ increases, the attainable accuracy degrades gradually, reaching $\mathrm{RMSE}\approx 4.7\times 10^{-2}$ at $\epsilon=0.05$.

\section{Conclusion}
We have presented a Wasserstein-inspired adversarial framework for unsupervised inference of sensor related transformation parameters, formulated as a distribution matching problem between reference and changed data regimes. By reinterpreting the generator as a deterministic transformation whose trainable weights correspond to physically meaningful parameters, the approach shifts the use of adversarial learning from sample synthesis to parameter recovery.
We validated the method in two complementary settings. In a simplified tracking toy model, the framework recovers injected geometric shifts in a low dimensional and fully controlled scenario, serving as a proof of concept for alignment style parameter inference. We then studied calorimeter aging using high granularity Geant4-simulated data with cell wise signal attenuation. In this setup, the inferred aging coefficients show good agreement with the injected values and, when applied as a calibration correction, improve the consistency between the calibrated and undamaged energy response distributions, without requiring labeled calibration targets or event-level correspondences.
The results also illustrate a key limitation of purely distribution based inference: as channel to channel stochastic variations increase and become comparable to the underlying systematic effect, the identifiability of individual parameters deteriorates due to reduced discriminative information in the observed distributions. This suggests that adversarial calibration is best viewed as a complementary technique that reduces reliance on supervision, rather than a replacement for established workflows. Future work will focus on extending the framework to time dependent degradation, more realistic detector geometries and readout effects, and hybrid strategies that incorporate conventional calibration constraints or external reference signals, with the aim of reducing recalibration overheads in long running experiments.

\section{Acknowledgments}
The work was supported by the grant for research centers in the field of AI provided by the Ministry of Economic Development of the Russian Federation in accordance with the agreement 000000C313925P4E0002 and the agreement with HSE University № 139-15-2025-009. The computation for this research was performed using the computational resources of HPC facilities at HSE University.

\bibliographystyle{unsrt}
\bibliography{references}

\end{document}